# Inequality Constraints in Causal Models with Hidden Variables

**Changsung Kang**
Department of Computer Science
Iowa State University
Ames, IA 50011
*cskang@iastate.edu*

**Jin Tian**
Department of Computer Science
Iowa State University
Ames, IA 50011
*jtian@cs.iastate.edu*

## Abstract

We present a class of inequality constraints on the set of distributions induced by local interventions on variables governed by a causal Bayesian network, in which some of the variables remain unmeasured. We derive bounds on causal effects that are not directly measured in randomized experiments. We derive instrumental inequality type of constraints on nonexperimental distributions. The results have applications in testing causal models with observational or experimental data.

## 1 Introduction

The use of graphical models for encoding distributional and causal information is now fairly standard [Heckerman and Shachter, 1995, Lauritzen, 2000, Pearl, 2000, Spirtes *et al.*, 2001]. The most common such representation involves a *causal Bayesian network (BN)*, namely, a directed acyclic graph (DAG) $G$ which, in addition to the usual conditional independence interpretation, is also given a causal interpretation. This additional feature permits one to infer the effects of interventions or actions, such as those encountered in policy analysis, treatment management, or planning. Specifically, if an external intervention fixes any set $T$ of variables to some constants $t$, the DAG permits us to infer the resulting post-intervention distribution, denoted by $P_t(v)$,[1] from the pre-intervention distribution $P(v)$. The quantity $P_t(y)$, often called the "causal effect" of $T$ on $Y$, is what we normally assess in a controlled experiment with $T$ randomized, in which the distribution of $Y$ is estimated for each level $t$ of $T$. We will call a post-intervention distribution an interventional distribution, and call the distribution $P(v)$ nonexperimental distribution.

[1][Pearl, 1995a, Pearl, 2000] used the notation $P(v|set(t))$, $P(v|do(t))$, or $P(v|\hat{t})$ for the post-intervention distribution, while [Lauritzen, 2000] used $P(v||t)$.

The validity of a causal model can be tested only if it has empirical implications, that is, it must impose constraints on the statistics of the data collected. A causal BN not only imposes constraints on the nonexperimental distribution but also on the interventional distributions that can be induced by the network. Therefore a causal BN can be tested and falsified by using two types of data, observational, which are passively observed, and experimental, which are produced by manipulating (randomly) some variables and observing the states of other variables. The ability to use a mixture of observational and experimental data will greatly increase our power of causal reasoning and learning. The use of a mixture of experimental and observational data in learning causal BN is demonstrated in [Cooper and Yoo, 1999, Heckerman, 1995]. In this paper we consider using combined observational and experimental data for causal model testing.

There has been much research on identifying observational implications of BNs. It is well-known that the observational implications of a BN are completely captured by conditional independence relationships among the variables when all the variables are observed [Pearl *et al.*, 1990]. When a BN invokes unobserved variables, called *hidden* or *latent* variables, the network structure may impose other equality and/or inequality constraints on the distribution of the observed variables [Verma and Pearl, 1990, Robins and Wasserman, 1997, Desjardins, 1999, Spirtes *et al.*, 2001]. Methods for identifying equality constraints were given in [Geiger and Meek, 1998, Tian and Pearl, 2002b]. [Pearl, 1995b] gave an example of inequality constraints in the model shown in Figure 1. The model imposes the following inequality, called the *instrumental inequality* by Pearl, for discrete variables $X, Y$, and $Z$,

$$\max_x \sum_y \max_z P(xy|z) \leq 1. \qquad (1)$$

This model has been further analysed using convex analysis approach in [Bonet, 2001]. In principle, all (equality and inequality) constraints implied by BNs with hidden variables can be derived by the quantifier elimination method presented in [Geiger and Meek, 1999]. However, due to high computational demand (doubly exponential in the number of probabilistic parameters), in practice, quantifier elimination is limited to BNs with few number of probabilistic parameters. For example, the current quantifier elimination algorithms cannot deal with the simple model in Figure 1 for $X$, $Y$, and $Z$ being binary variables.

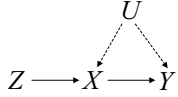

Figure 1: $U$ is a hidden variable.

When all variables are observed, a complete characterization of constraints on interventional distributions imposed by a given causal BN has been given in [Pearl, 2000, pp.23-4]. When a causal BN contains unobserved variables, there may be inequality constraints on interventional distributions [Tian and Pearl, 2002a]. For the model in Figure 1, bounds on causal effects $P_x(y)$ in terms of the nonexperimental distribution $P(x, y, z)$ was derived in [Balke and Pearl, 1994, Chickering and Pearl, 1996] using linear programming method for $X$, $Y$, and $Z$ being binary variables.

In this paper, we seek the constraints imposed by a causal BN with hidden variables on both nonexperimental and interventional distributions. We present a type of inequality constraints on interventional distributions. We derive bounds on causal effects in terms of nonexperimental distributions and given interventional distributions. We derive instrumental inequality type of constraints upon nonexperimental distributions. Although the constraints we give are not complete, they constitute necessary conditions for a hypothesized model to be compatible with the data. The constraints also provide information (bounds) on the effects of interventions that have not been tried experimentally, from observational data and given experimental data.

## 2 Causal Bayesian Networks and Interventions

A causal Bayesian network, also known as a *Markovian model*, consists of two mathematical objects: (i) a DAG $G$, called a *causal graph*, over a set $V = \{V_1, \ldots, V_n\}$ of vertices, and (ii) a probability distribution $P(v)$, over the set $V$ of discrete variables that correspond to the vertices in $G$.[2] The interpretation of such a graph has two components, probabilistic and causal. The probabilistic interpretation views $G$ as representing conditional independence restrictions on $P$: Each variable is independent of all its non-descendants given its direct parents in the graph. These restrictions imply that the joint probability function $P(v) = P(v_1, \ldots, v_n)$ factorizes according to the product

$$P(v) = \prod_i P(v_i|pa_i) \quad (2)$$

where $pa_i$ are (values of) the parents of variable $V_i$ in $G$.

The causal interpretation views the arrows in $G$ as representing causal influences between the corresponding variables. In this interpretation, the factorization of (2) still holds, but the factors are further assumed to represent autonomous data-generation processes, that is, each conditional probability $P(v_i|pa_i)$ represents a stochastic process by which the values of $V_i$ are assigned in response to the values $pa_i$ (previously chosen for $V_i$'s parents), and the stochastic variation of this assignment is assumed independent of the variations in all other assignments in the model. Moreover, each assignment process remains invariant to possible changes in the assignment processes that govern other variables in the system. This modularity assumption enables us to predict the effects of interventions, whenever interventions are described as specific modifications of some factors in the product of (2). The simplest such intervention, called *atomic*, involves fixing a set $T$ of variables to some constants $T = t$, which yields the post-intervention distribution

$$P_t(v) = \begin{cases} \prod_{\{i|V_i \notin T\}} P(v_i|pa_i) & v \text{ consistent with } t. \\ 0 & v \text{ inconsistent with } t. \end{cases} \quad (3)$$

Eq. (3) represents a truncated factorization of (2), with factors corresponding to the manipulated variables removed. This truncation follows immediately from (2) since, assuming modularity, the post-intervention probabilities $P(v_i|pa_i)$ corresponding to variables in $T$ are either 1 or 0, while those corresponding to unmanipulated variables remain unaltered. If $T$ stands for a set of treatment variables and $Y$ for an outcome variable in $V \setminus T$, then Eq. (3) permits us to calculate the probability $P_t(y)$ that event $Y = y$ would occur if treatment condition $T = t$ were enforced uniformly over the population.

When some variables in a Markovian model are unobserved, the probability distribution over the observed variables may no longer be decomposed as in Eq. (2). Let $V = \{V_1, \ldots, V_n\}$ and $U = \{U_1, \ldots, U_{n'}\}$ stand for the sets of observed and unobserved variables respectively. If no $U$ variable is a descendant of any $V$ variable, then the corresponding model is called a *semi-Markovian model*. In this paper, we only consider semi-Markovian models. However, the results can be generalized to models with arbitrary unobserved variables as shown in [Tian and Pearl, 2002b]. In a semi-Markovian model, the observed probability distribution, $P(v)$, becomes a mixture of products:

$$P(v) = \sum_u \prod_i P(v_i|pa_i, u^i)P(u) \quad (4)$$

---

[2] We only consider discrete random variables in this paper.

where $PA_i$ and $U^i$ stand for the sets of the observed and unobserved parents of $V_i$, and the summation ranges over all the $U$ variables. The post-intervention distribution, likewise, will be given as a mixture of truncated products

$$P_t(v) = \begin{cases} \sum_u \prod_{\{i|V_i \notin T\}} P(v_i|pa_i, u^i)P(u) & v \text{ consistent with } t. \\ 0 & v \text{ inconsistent with } t. \end{cases} \quad (5)$$

Assuming that $v$ is consistent with $t$, we can write

$$P_t(v) = P_t(v \setminus t) \quad (6)$$

In the rest of the paper, we will use $P_t(v)$ and $P_t(v \setminus t)$ interchangeably, always assuming $v$ being consistent with $t$.

## 3 Constraints on Interventional Distributions

Let $\boldsymbol{P_*}$ denote the set of all interventional distributions induced by a given semi-Markovian model,

$$\boldsymbol{P_*} = \{P_t(v) | T \subseteq V, t \in Dm(T), v \in Dm(V)\} \quad (7)$$

where $Dm(T)$ represents the domain of $T$. What are the constraints imposed by the model on the interventional distributions in $\boldsymbol{P_*}$? The structure of the causal graph $G$ will play an important role in finding these constraints. A *c-component* is a maximal set of vertices such that any two vertices in the set are connected by a path on which every edge is of the form $\leftarrow\text{-- } U \text{ --}\rightarrow$ where $U$ is a hidden variable. The set of variables $V$ is then partitioned into a set of c-components. For example, the causal graph $G$ in Figure 2 consists of two c-components $\{X, Y, Z\}$ and $\{W_1, W_2\}$.

Let $G(H)$ denote the subgraph of $G$ composed only of the variables in $H$ and the hidden variables that are ancestors of $H$. In general, equality constraints on the set of interventional distributions can be found using the following three lemmas.

**Lemma 1** [Tian and Pearl, 2002b] *Let $H \subseteq V$, and assume that $H$ is partitioned into c-components $H_1, \ldots, H_l$ in the subgraph $G(H)$. Then we have*

(i) $P_{v \setminus h}(v)$ *decomposes as*

$$P_{v \setminus h}(v) = \prod_i P_{v \setminus h_i}(v). \quad (8)$$

(ii) *Let $k$ be the number of variables in $H$, and let a topological order of the variables in $H$ be $V_{h_1} < \ldots < V_{h_k}$ in $G(H)$. Let $H^{(i)} = \{V_{h_1}, \ldots, V_{h_i}\}$ be the set of variables in $H$ ordered before $V_{h_i}$ (including $V_{h_i}$), $i = 1, \ldots, k$, and $H^{(0)} = \emptyset$. Then each $P_{v \setminus h_j}(v)$,*

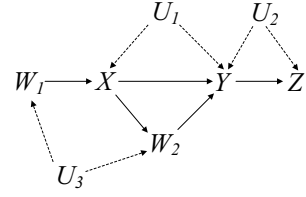

Figure 2: $U_1, U_2$ and $U_3$ are hidden variables.

$j = 1, \ldots, l$, *is computable from $P_{v \setminus h}(v)$ and is given by*

$$P_{v \setminus h_j}(v) = \prod_{\{i|V_{h_i} \in H_j\}} \frac{P_{v \setminus h^{(i)}}(v)}{P_{v \setminus h^{(i-1)}}(v)}, \quad (9)$$

*where each $P_{v \setminus h^{(i)}}(v)$, $i = 0, 1, \ldots, k$, is given by*

$$P_{v \setminus h^{(i)}}(v) = \sum_{h \setminus h^{(i)}} P_{v \setminus h}(v). \quad (10)$$

A special case of Lemma 1 is when $H = V$, and we have the following Lemma.

**Lemma 2** [Tian and Pearl, 2002b] *Assuming that $V$ is partitioned into c-components $S_1, \ldots, S_k$, we have*

(i) $P(v) = \prod_i P_{v \setminus s_i}(v)$.

(ii) *Let a topological order over $V$ be $V_1 < \ldots < V_n$, and let $V^{(i)} = \{V_1, \ldots, V_i\}, i = 1, \ldots, n$, and $V^{(0)} = \emptyset$. Then each $P_{v \setminus s_j}(v), j = 1, \ldots, k$, is computable from $P(v)$ and is given by*

$$P_{v \setminus s_j}(v) = \prod_{\{i|V_i \in S_j\}} P(v_i|v^{(i-1)}). \quad (11)$$

The next lemma provides a condition under which we can compute $P_{v \setminus w}(w)$ from $P_{v \setminus c}(c)$ where $W$ is a subset of $C$, by simply summing $P_{v \setminus c}(c)$ over other variables $C \setminus W$.

**Lemma 3** [Tian and Pearl, 2002b] *Let $W \subseteq C \subseteq V$, and $W' = C \setminus W$. If $W$ contains its own observed ancestors in $G(C)$, then*

$$\sum_{w'} P_{v \setminus c}(v) = P_{v \setminus w}(v). \quad (12)$$

The set of equality constraints implied by these three lemmas can be systematically listed by slightly modifying the procedure in [Tian and Pearl, 2002b] for listing equality constraints on nonexperimental distributions. We will not show the details of the procedure here since the focus of this paper is on inequality constraints.

For example, the model in Figure 1 imposes the following equality constraints.

$$P_z(xy) = P(xy|z) \qquad (13)$$
$$P_{yz}(x) = P(x|z) \qquad (14)$$
$$P_{xz}(y) = P_x(y) \qquad (15)$$

The model in Figure 2 imposes the following equality constraints.

$$P_{w_1w_2}(xyz) = P(z|w_1xw_2y)P(y|w_1xw_2)P(x|w_1) \qquad (16)$$

$$P_{w_1w_2z}(xy) = P(y|w_1xw_2)P(x|w_1) \qquad (17)$$
$$P_{w_1w_2y}(xz) = P_{w_1y}(xz) \qquad (18)$$
$$P_{w_1w_2x}(yz) = P_{w_2x}(yz) \qquad (19)$$
$$P_{w_1w_2yz}(x) = P(x|w_1) \qquad (20)$$
$$P_{w_1w_2xz}(y) = P_{w_2x}(y) \qquad (21)$$
$$P_{w_1w_2xy}(z) = P_y(z) \qquad (22)$$
$$P_{xyz}(w_1w_2) = P(w_2|w_1x)P(w_1) \qquad (23)$$
$$P_{xyzw_2}(w_1) = P(w_1) \qquad (24)$$
$$P_{xyzw_1}(w_2) = \sum_{w_1} P(w_2|w_1x)P(w_1) \qquad (25)$$

### 3.1 Inequality Constraints

In this paper, we are concerned with inequality constraints imposed by a model. The $P_*$ set induced from a semi-Markovian model must satisfy the following inequality constraints.

**Lemma 4** *For any $S_1 \subseteq V$ and any superset $S_1' \subseteq V$ of $S_1$, we have*

$$\sum_{S_2 \subseteq S_1' \setminus S_1} (-1)^{|S_2|} P_{v \setminus (s_1 \cup s_2)}(v) \geq 0, \quad \forall v \in Dm(V) \qquad (26)$$

*where $|S_2|$ represents the number of variables in $S_2$.*

**Proof:** We use the following equation.

$$\prod_{i=1}^{k}(1 - a_i)$$
$$= 1 - \sum_i a_i + \sum_{i,j} a_i a_j - \ldots + (-1)^k a_1 \ldots a_k. \qquad (27)$$

Take $a_j = P(v_j|pa_j, u^j)$, we have that

$$\sum_u \prod_{\{i|V_i \in S_1\}} P(v_i|pa_i, u^i)$$
$$\prod_{\{j|V_j \in S_1' \setminus S_1\}} (1 - P(v_j|pa_j, u^j))P(u)$$
$$= \sum_{S_2 \subseteq S_1' \setminus S_1} (-1)^{|S_2|} P_{v \setminus (s_1 \cup s_2)}(v) \geq 0 \qquad (28)$$

since for all $V_i \in V$

$$0 \leq P(v_i|pa_i, u^i) \leq 1. \qquad (29)$$

■

For a fixed $S_1'$ set, there are $2^{|S_1'|}$ number of Eq. (26) type of inequalities. For different $S_1'$ sets, some of those inequalities may imply others as shown in the following lemma.

**Lemma 5** *If $S_1' \subset S_1''$, then the set of $2^{|S_1''|}$ inequalities, $\forall S_1 \subseteq S_1''$,*

$$\sum_{S_2 \subseteq S_1'' \setminus S_1} (-1)^{|S_2|} P_{v \setminus (s_1 \cup s_2)}(v) \geq 0, \quad \forall v \in Dm(V) \qquad (30)$$

*imply the set of $2^{|S_1'|}$ inequalities, $\forall S_1 \subseteq S_1'$,*

$$\sum_{S_2 \subseteq S_1' \setminus S_1} (-1)^{|S_2|} P_{v \setminus (s_1 \cup s_2)}(v) \geq 0, \quad \forall v \in Dm(V) \qquad (31)$$

The proof is omitted due to space limitation.

Assume that the set of variables $V$ in the model is partitioned into c-components $T_1, \ldots, T_k$. Due to the equality constraints given in Lemma 1, instead of listing $2^{|V|}$ Eq. (26) type of inequalities, we only need to give $2^{|T_i|}$ Eq. (26) type of inequalities for each c-component $T_i$.

**Proposition 1** *Let the set of variables $V$ in a semi-Markovian model be partitioned into c-components $T_1, \ldots, T_k$. The $P_*$ set must satisfy the following inequality constraints: for $i = 1, \ldots, k$, $\forall S_1 \subseteq T_i$,*

$$\sum_{S_2 \subseteq T_i \setminus S_1} (-1)^{|S_2|} P_{v \setminus (s_1 \cup s_2)}(v) \geq 0, \quad \forall v \in Dm(V) \qquad (32)$$

For example, Proposition 1 gives the following inequality constraints for the model in Figure 1,

$$1 - P_{yz}(x) - P_{xz}(y) + P_z(xy) \geq 0 \qquad (33)$$
$$P_{yz}(x) - P_z(xy) \geq 0 \qquad (34)$$
$$P_{xz}(y) - P_z(xy) \geq 0 \qquad (35)$$
$$P_z(xy) \geq 0, \qquad (36)$$

in which (36) is trivial, and (34) becomes trivial because of equality constraints (13) and (14).

For the model in Figure 2, Proposition 1 gives the following inequality constraints for the c-component $\{X, Y, Z\}$,

$$1 - P_{w_1 w_2 yz}(x) - P_{w_1 w_2 xz}(y) - P_{w_1 w_2 xy}(z)$$
$$+ P_{w_1 w_2 z}(xy) + P_{w_1 w_2 y}(xz) + P_{w_1 w_2 x}(yz)$$
$$- P_{w_1 w_2}(xyz) \geq 0 \quad (37)$$

$$P_{w_1 w_2 yz}(x) - P_{w_1 w_2 z}(xy) - P_{w_1 w_2 y}(xz)$$
$$+ P_{w_1 w_2}(xyz) \geq 0 \quad (38)$$

$$P_{w_1 w_2 xz}(y) - P_{w_1 w_2 z}(xy) - P_{w_1 w_2 x}(yz)$$
$$+ P_{w_1 w_2}(xyz) \geq 0 \quad (39)$$

$$P_{w_1 w_2 xy}(z) - P_{w_1 w_2 y}(xz) - P_{w_1 w_2 x}(yz)$$
$$+ P_{w_1 w_2}(xyz) \geq 0 \quad (40)$$

$$P_{w_1 w_2 z}(xy) - P_{w_1 w_2}(xyz) \geq 0 \quad (41)$$
$$P_{w_1 w_2 y}(xz) - P_{w_1 w_2}(xyz) \geq 0 \quad (42)$$
$$P_{w_1 w_2 x}(yz) - P_{w_1 w_2}(xyz) \geq 0 \quad (43)$$
$$P_{w_1 w_2}(xyz) \geq 0, \quad (44)$$

some of which are implied by the set of equality constraints (16)-(25). It can be shown that all inequality constraints for c-component $\{W_1, W_2\}$ are implied by equality constraints.

Note that in general, the inequality constraints given in this section are not the complete set of constraints that are implied by a given model. For example, for the model given in Figure 1, the sharp bounds on $P_x(y)$ given in [Balke and Pearl, 1994] for $X$, $Y$, and $Z$ being binary variables are not implied by (33)-(36).

## 4 Inequality Constraints On a Subset of Interventional Distributions

Proposition 1 gives a set of inequality constraints on the set of interventional distributions in $\boldsymbol{P_*}$. In practical situations, we may be interested in constraints involving only a certain subset of interventional distributions. For example, (i) We have done some experiments, and obtained $P_s(v)$ for some sets $S$. We want to know whether these data are compatible with the given model. For this purpose, we would like inequality constraints involving only those known interventional distributions; (ii) A special case of (i) is that we only have the nonexperimental distribution $P(v)$. We want inequality constraints on $P(v)$ imposed by the model; (iii) In practice, certain experiments may be difficult or expensive to perform. Still, we want some information about a particular causal effect, given some known interventional distributions and nonexperimental distribution. We may provide bounds on concerned causal effect that can be derived from those inequality constraints (if this causal effect is not computable from given quantity through equality constraints).

To restrict the set of inequality constraints given in Proposition 1 to constraints involving only certain subset of interventional distributions, in principle, we can treat each $P_s(v)$ for an instantiation of $v \in Dm(V)$ as a variable, and solve the inequalities to eliminate unwanted variables using methods like Fourier-Motzkin elimination or quantifier elimination. However, this is typically only practical for small number of binary variables due to high computational complexity. In this paper, we show some inequality constraints involving only interventional distributions of interests that can be derived from those in Proposition 1. In general, these constraints may not include all the possible constraints that could be derived from Proposition 1 in principle.

Instead of directly solving the inequality constraints given in Proposition 1, we consider the inequality in Eq. (26) for every $S_1' \subseteq T_i$. We keep every inequality that involves only the interventional distributions of interests. Those inequalities that contain unwanted interventional distributions may lead to some new inequalities. For example, in the model in Figure 2, consider the following inequality that follows from (26) with $S_1 = \{Z\}$ and $S_1' = \{Y, Z\}$,

$$P_{w_1 w_2 xy}(z) - P_{w_1 w_2 x}(yz) \geq 0. \quad (45)$$

Suppose we want constraints on $P_{w_1 w_2 x}(yz)$ and get rid of unknown quantity $P_{w_1 w_2 xy}(z)$. First we have equality constraints (19) and (22), and Eq. (45) becomes

$$P_{w_2 x}(yz) \leq P_y(z) \quad (46)$$

$P_{w_2 x}(yz)$ is a function of $W_2$ and $X$ but $P_y(z)$ is not, which leads to

$$\max_{w_2, x} P_{w_2 x}(yz) \leq P_y(z) \quad (47)$$

$$\sum_z \max_{w_2, x} P_{w_2 x}(yz) \leq 1 \quad (48)$$

Eq. (48) is a nontrivial inequality constraint on $P_{w_1 w_2 x}(yz) = P_{w_2 x}(yz)$, which can also be represented as

$$P_{w_2 x}(yz_0) + P_{w_2' x'}(yz_1) \leq 1 \quad (49)$$

for any $w_2 \in Dm(W_2)$, $x \in Dm(X)$, $w_2' \in Dm(W_2)$ and $x' \in Dm(X)$ when $Z$ is binary ($Dm(Z) = \{z_0, z_1\}$).

From the above considerations, we give a procedure in Figure 3 that lists the inequality constraints on the interventional distributions of interest. The procedure has a complexity of $3^{2|T_i|}$. Note that $A$ will always contain the nonexperimental distribution and all interventional distributions that can be computed from $P(v)$ (via equality constraints).

In Step 1, we list the inequalities that do not involve unwanted quantities (i.e., interventional distributions not included in $A$). Note that we remove some redundant inequalities based on the following lemma.

**procedure FindIneqs**(G,A)
**INPUT:** a causal graph $G$, interventional distributions of interest $A$, equality constraints implied by $G$
**OUTPUT:** inequalities of interests, $IE_{T_i}$ for each c-component $T_i, i = 1, \ldots, k$
**Step 1:**
**For each** c-component $T_i, i = 1, \ldots, k$
  **For each** $S_1 \subseteq T_i$ (small to large)
    **For each** $S_1' \subseteq T_i$ such that $S_1 \subseteq S_1'$ (small to large)
    Study the inequality
    $e_{S_1,S_1'} = \sum_{S_2 \subseteq S_1' \setminus S_1} (-1)^{|S_2|} P_{v \setminus (s_1 \cup s_2)}(v) \geq 0$
    **If** every interventional distribution in $e_{S_1,S_1'}$ is in $A$
      $IE_{T_i} = IE_{T_i} \cup \{e_{S_1,S_1'} \geq 0\}$;
      Remove any $e_{S_1,R}$ in $IE_{T_i}$ such that $R \subset S_1'$;
**Step 2:**
**For each** c-component $T_i, i = 1, \ldots, k$
  **For each** $S_1 \subseteq T_i$ (small to large)
    **For each** $S_1' \subseteq T_i$ such that $S_1 \subseteq S_1'$ (small to large)
    Study the inequality
    $e_{S_1,S_1'} = \sum_{S_2 \subseteq S_1' \setminus S_1} (-1)^{|S_2|} P_{v \setminus (s_1 \cup s_2)}(v) \geq 0$
    **If** some interventional distribution in $e_{S_1,S_1'}$ is not in $A$
      $IE_{T_i} = IE_{T_i} \cup \{e_{S_1,S_1'} \geq 0$ reformulated in the form of (55)$\}$;

Figure 3: A Procedure for Listing Inequality Constraints On a Subset of Interventional Distributions

**Lemma 6** *Let $Sup(S_1)$ denote the set of supersets of $S_1$ such that $S_1' \in Sup(S_1)$ if and only if every interventional distribution in $e_{S_1,S_1'} = \sum_{S_2 \subseteq S_1' \setminus S_1} (-1)^{|S_2|} P_{v \setminus (s_1 \cup s_2)}(v) \geq 0$ is in $A$. For a set of sets $W$, let $Max(W) = \{S | S \in W, \text{there is no } S' \in W \text{ such that } S \subset S'\}$ denote the set of maximal sets in $W$. Then, the set of inequalities*

$$\forall S_1 \subseteq T_i, \forall S_1' \in Max(Sup(S_1)),$$
$$\sum_{S_2 \subseteq S_1' \setminus S_1} (-1)^{|S_2|} P_{v \setminus (s_1 \cup s_2)}(v) \geq 0, \forall v \in Dm(V) \quad (50)$$

*imply the inequalities*

$$\forall S_1 \subseteq T_i, \forall S_1' \in Sup(S_1)$$
$$\sum_{S_2 \subseteq S_1' \setminus S_1} (-1)^{|S_2|} P_{v \setminus (s_1 \cup s_2)}(v) \geq 0, \forall v \in Dm(V). \quad (51)$$

See the Appendix for the proof.

In Step 2, we deal with the inequalities that contain unwanted quantities as follows. We rewrite the inequality in Eq. (26) as $e_{S_1,S_1'} \geq 0$, with

$$e_{S_1,S_1'}$$
$$= \sum_{R \in W_1} (-1)^{|R|-|S_1|} P_{v \setminus r}(v) + \sum_{R \in W_2} (-1)^{|R|-|S_1|} P_{v \setminus r}(v) \quad (52)$$

where $W_1 = \{S_1 \cup S_2 | S_2 \subseteq S_1' \setminus S_1, P_{v \setminus (s_1 \cup s_2)}(v)$ is in $A\}$ and $W_2 = \{S_1 \cup S_2 | S_2 \subseteq S_1' \setminus S_1, P_{v \setminus (s_1 \cup s_2)}(v)$ is not in $A\}$. We have

$$\sum_{R \in W_1} (-1)^{|R|-|S_1|} P_{v \setminus r}(v) \geq -\sum_{R \in W_2} (-1)^{|R|-|S_1|} P_{v \setminus r}(v). \quad (53)$$

Suppose the left-hand side is a function of variables $E_1$ and the right-hand side is a function of variables $E_2$. Then,

$$\min_{E_1 \setminus E_2} \sum_{R \in W_1} (-1)^{|R|-|S_1|} P_{v \setminus r}(v)$$
$$\geq -\sum_{R \in W_2} (-1)^{|R|-|S_1|} P_{v \setminus r}(r). \quad (54)$$

Let $Q = \bigcup_{R \in W_2} R$. We obtain,

$$\sum_Q \min_{E_1 \setminus E_2} \sum_{R \in W_1} (-1)^{|R|-|S_1|} P_{v \setminus r}(v)$$
$$\geq -\sum_{R \in W_2} (-1)^{|R|-|S_1|} \prod_{\{i | V_i \in Q \setminus R\}} |Dm(V_i)|. \quad (55)$$

Note that if $E_1 \setminus E_2 = \emptyset$, then we do not need $\min_{E_1 \setminus E_2}$.

To illustrate the procedure, suppose we want to get the inequality constraints on the two interventional distributions $P_{w_1 w_2 xy}(z)$ and $P_{w_1 w_2 x}(yz)$ and we are given a tried interventional distribution $P_{w_1 w_2 y}(xz)$.

In Step 1, consider the loop in which $T_i = \{X, Y, Z\}$ and $S_1 = \{\emptyset\}$. The procedure first adds $e_{\emptyset, \{X\}}$ and $e_{\emptyset, \{Z\}}$. When it adds $e_{\emptyset, \{X,Z\}}$, it will remove $e_{\emptyset, \{X\}}$ and $e_{\emptyset, \{Z\}}$ from $IE_{T_i}$ and keep $e_{\emptyset, \{X,Z\}}$ which turns out to be $Max(Sup(\emptyset))$. This repeats for every $S_1 \subseteq T_i$.

In Step 2, consider the loop where $T_i = \{X, Y, Z\}$ and $S_1 = \{Y\}$. The procedure studies $e_{S_1,S_1'}$ for each $S_1' \in \{\{Y\}, \{X,Y\}, \{Y,Z\}, \{X,Y,Z\}\}$. For example, for $S_1' = \{X, Y, Z\}$, we have the inequality (39). From (16), (17), (19) and (21), we obtain

$$\max_{w_1, z} \Big( P(y|w_1 x w_2) P(x|w_1) + P_{w_2 x}(yz)$$
$$- P(z|w_1 x w_2 y) P(y|w_1 x w_2) P(x|w_1) \Big) \leq P_{w_2 x}(y). \quad (56)$$

Summing both sides over $Y$ gives

$$\sum_y \max_{w_1, z} \Big( P(y|w_1 x w_2) P(x|w_1) + P_{w_2 x}(yz)$$
$$- P(z|w_1 x w_2 y) P(y|w_1 x w_2) P(x|w_1) \Big) \leq 1. \quad (57)$$

### 4.1 Bounds on Causal Effects

Suppose that our goal is to bound a particular interventional distribution. For this case, $A$ in the procedure **FindIneqs** consists of the particular interventional distribution that we want to bound, the nonexperimental distribution $P(v)$, and all interventional distributions that are computable from $P(v)$.

For example, consider the graph in Figure 2. Suppose that we want to bound the interventional distribution $P_{w_1 w_2 xy}(z)$ and that the interventional distribution $P_{w_1 w_2 y}(xz)$ is available from experiments. **FindIneqs** will list the following bounds for $P_{w_1 w_2 xy}(z)$ in Step 1.

$$1 - P(x|w_1) - P_{w_1 w_2 xy}(z) + P_{w_1 w_2 y}(xz) \geq 0 \quad (58)$$
$$P_{w_1 w_2 xy}(z) - P_{w_1 w_2 y}(xz) \geq 0 \quad (59)$$

which provides a lower and upper bound for $P_{w_1 w_2 xy}(z)$ respectively.

### 4.2 Inequality Constraints on Nonexperimental Distribution

Now assume that we want to find inequality constraints on nonexperimental distribution. For this case, $A$ in the procedure **FindIneqs** consists of the nonexperimental distribution $P(v)$ and all interventional distributions that are computable from $P(v)$.

The inequality constraints produced by **FindIneqs** in this case include the instrumental inequality type of constraints. Consider the graph in Figure 1. For the c-component $\{X, Y\}$, **FindIneqs** will produce the inequality (35). From (13) and (15), we have

$$\max_z P(xy|z) \leq P_x(y) \quad (60)$$

and summing both sides over $Y$ gives

$$\sum_y \max_z P(xy|z) \leq 1. \quad (61)$$

Since this must hold for all $X$, we obtain the instrumental inequality (1).

To illustrate more general instrumental inequality type of constraints, consider the graph in Figure 2. For $S_1 = \{Y, Z\}$ and $S_1' = \{X, Y, Z\}$, **FindIneqs** produces the inequality (43). From (16) and (19), we have

$$\max_{w_1} P(z|w_1 x w_2 y) P(y|w_1 x w_2) P(x|w_1) \leq P_{w_2 x}(yz). \quad (62)$$

Summing both sides over $Y$ and $Z$ gives

$$\sum_{yz} \max_{w_1} P(z|w_1 x w_2 y) P(y|w_1 x w_2) P(x|w_1) \leq 1. \quad (63)$$

## 5 Conclusion

We present a class of inequality constraints imposed by a given causal BN with hidden variables on the set of interventional distributions that can be induced from the network. We show a method to restrict these inequality constraints on to that only involving interventional distributions of interests. These inequality constraints can be used as necessary test for a causal model to be compatible with given observational and experimental data. Another application permits us to bound the effects of untried interventions from experiments involving auxiliary interventions that are easier or cheaper to implement.

We derive a type of inequality constraints upon the nonexperimental distribution in a complexity of $3^{2m}$ where $m$ is the number of variables in the largest c-component. These constraints are imposed by the network structure, regardless of the number of states of the (observed or hidden) variables involved. These constraints can be used to test a model or distinguish between models. How to test these inequality constraints in practice and use them for model selection would be interesting future research.

### Acknowledgments

This research was partly supported by NSF grant IIS-0347846.

## Appendix : Proof of Lemma 6

We will show that if the inequalities in (50) hold, then for any $n \leq |V|$ we have

$$\forall S_1 \subseteq T_i, \forall S_1' \in Max^n(Sup(S_1)),$$
$$\sum_{S_2 \subseteq S_1' \setminus S_1} (-1)^{|S_2|} P_{v \setminus (s_1 \cup s_2)}(v) \geq 0, \forall v \in Dm(V) \quad (64)$$

where $Max^n(S) = Max(S \setminus \{R|R \in S, |R| > n\})$. (51) will follow from (64) if we let $n$ be the size of the set $S_1'$ in (51). Assuming (50), we prove (64) by induction on $n$.

Base: $n = |V|$. (64) is equivalent to (50).

Hypothesis: Assume that

$$\forall S_1 \subseteq T_i, \forall S_1' \in Max^n(Sup(S_1)),$$
$$\sum_{S_2 \subseteq S_1' \setminus S_1} (-1)^{|S_2|} P_{v \setminus (s_1 \cup s_2)}(v) \geq 0, \forall v \in Dm(V). \quad (65)$$

Induction step: We show that

$$\forall S_1 \subseteq T_i, \forall S_1' \in Max^{n-1}(Sup(S_1)),$$
$$\sum_{S_2 \subseteq S_1' \setminus S_1} (-1)^{|S_2|} P_{v \setminus (s_1 \cup s_2)}(v) \geq 0, \forall v \in Dm(V). \quad (66)$$

If $|S_1'| < n-1$, then $S_1'$ is in $Max^n(Sup(S_1))$. Thus, (66) follows from (65). If $|S_1'| = n-1$, then one of the followings should hold.

Case 1: $S_1'$ is in $Max^n(Sup(S_1))$.

Case 2: There exists a variable $\alpha$ such that $S_1' \cup \{\alpha\}$ is in $Max^n(Sup(S_1))$.

In Case 1, (66) follows from (65). In Case 2, we have

$$\sum_{S_2 \subseteq (S_1' \cup \{\alpha\}) \setminus S_1} (-1)^{|S_2|} P_{v \setminus (s_1 \cup s_2)}(v) \geq 0, \forall v \in Dm(V) \tag{67}$$

and

$$\sum_{S_2 \subseteq S_1' \setminus S_1} (-1)^{|S_2|} P_{v \setminus (s_1 \cup \{\alpha\} \cup s_2)}(v) \geq 0, \forall v \in Dm(V). \tag{68}$$

(68) follows from (65) since $S_1' \cup \{\alpha\}$ is in $Max^n(Sup(S_1 \cup \{\alpha\}))$. Summing (67) and (68) gives (66). ■